\documentclass[pmlr,twocolumn,a4paper,10pt]{jmlr}
\usepackage{isipta2023}
\usepackage[american]{babel}
\usepackage{booktabs} 
\usepackage{tikz} 

\newcommand{\Pcal}{\mathcal{P}}
\newcommand{\Sscr}{\mathscr{S}}
\newcommand{\Cscr}{\mathscr{C}}

\jmlrworkshop{ISIPTA 2023} 

\title[Depth Functions for Partial Orders]{
 Depth Functions for Partial Orders\\ with a Descriptive Analysis of Machine Learning Algorithms\titlebreak~ 
}

\author{
  \Name{Hannah Blocher}\Email{hannah.blocher@stat.uni-muenchen.de}\\ 
  \Name{Georg Schollmeyer}\Email{georg.schollmeyer@stat.uni-muenchen.de}\\
  \Name{Christoph Jansen}\Email{christoph.jansen@stat.uni-muenchen.de}\\
  \Name{Malte Nalenz}\Email{malte.nalenz@stat.uni-muenchen.de}\\
  \addr Department of Statistics, Ludiwg--Maximilians--Universität München, Munich, Germany
}

\begin{document}
\maketitle

\begin{abstract}
We propose a framework for descriptively analyzing sets of partial orders based on the concept of depth functions. 
Despite intensive studies of depth functions in linear and metric spaces, there is very little discussion on depth functions for non-standard data types such as partial orders.
We introduce an adaptation of the well-known simplicial
depth to the set of all partial orders, the union-free generic (ufg) depth. 
Moreover, we utilize our ufg depth for a comparison of machine learning algorithms based on multidimensional performance measures. Concretely, we analyze the distribution of
different classifier performances over a sample of standard benchmark data sets. Our results promisingly demonstrate that our approach differs substantially from existing benchmarking approaches and, therefore, adds a new perspective to the vivid debate on the comparison of classifiers.\footnote{\textbf{Open Science:} Reproducible implementation and data analysis are available at: \url{www.github.com/hannahblo/23_Performance_Analysis_ML_Algorithms}.}



\vspace{1em}
\end{abstract}
\begin{keywords} partial orders, data depth, benchmarking, algorithm comparison, outlier detection, non-standard data
\end{keywords}

\section{Introduction}
  \textit{Partial orders} -- and the systematic incomparabilities of objects encoded in them -- occur naturally in a variety of problems in a wide range of scientific disciplines. Examples range from decision theory, where the agents under consideration might be unable to arrange the consequences of their actions into total orders (see, e.g.,~\cite{sks1995,kikuti}) or have partial cardinal preferences (see, e.g.,~\cite{jsa2018,jbas2022}), over  social choice theory, where a fair aggregate order might only be possible by incorporating systematic incomparabilities (see, e.g.,~\cite{p2012,jsa2018b}), to finance, where risky assets do not always have to be comparable (see, e.g.,~\cite{ll1984,c2015}). Of course, many other relevant examples exist.
  
In the specific context of statistics and machine learning, the incompleteness of the considered orders often originates from the fact that the objects to be ordered are to be compared with respect to several criteria and/or on several instances \textit{simultaneously}: only if there is unanimous dominance of one object over another, this order is included in the corresponding relation. 
Quite a number of research papers recently have been devoted to such comparison in the specific context of classification algorithms, either with respect to multiple quality metrics~(e.g.,~\cite{ehl2012,jnsa2022}) or across multiple data sets (e.g.,~\cite{d2006,bcm2016}) or with respect to genuinely multidimensional performance criteria like receiver operating characteristic (ROC) curves (e.g.,~\cite{c2020}). Another source of partial incomparability of classifiers is the case  of classifiers that make only imprecise predictions, like for example the naive credal classifier (cf., \cite{ZAFFALON20025}) or credal sum-product networks (cf., \cite{pmlr-v62-maua17a}). In this case the imprecision in the predictions may take over to incomparabilities of the then possibly interval-valued performance measures\footnote{For example one could think of comparing classifiers with  utility-discounted predictive accuracy, cf., \cite[p. 1292 ]{ZAFFALON20121282} under the usage of a whole range $[\underline{a},\overline{a}]$ for the coefficient of risk aversion.}
 
Within the application field of machine learning and statistics, one further aspect is of special importance: Since the instances generally depend on chance, the same is true for the partial orders considered. Consequently, instead of a single partial order, random variables must then be analyzed that map into the set of \textit{all possible} partial orders on the set of objects under consideration. For example, in the aforementioned comparison of classification algorithms, the concrete order obtained depends on the random instantiation of the data set on which they are applied. In this paper, we are interested in exactly this situation: we discuss ways to \textit{descriptively} analyze samples of such partial order-valued (or short: \textit{poset-valued}\footnote{Note that in fact we speak here about random variables which have posets (on a common underlying ground space) as outcomes. This should not be confused with random  variables which have values in a partially ordered set.}) random variables.

Before starting, it is worth taking the time to distinguish, right at the beginning, those works that we believe are closest to ours. The main difference from the analysis in~\cite{jnsa2022}, which addresses a similar setting, is that in our case, in addition to the emphasis on the descriptive rather than the inferential aspects, the random orders do not (necessarily) arise retrospectively from the pairwise comparisons of the individual algorithms. Rather, they are conceived as abstract random objects. Further, the goal of our paper is also very different: while~\cite{jnsa2022} is interested in exploiting the available information in the best possible way to give one \textit{global} partial order over the classifiers under consideration, the present paper aims to analyze the \textit{distribution} of the partial classifier orders over a given set of data sets. The two methods are therefore -- despite similarity of the formal setting -- very different and thus not directly comparable.

Additionally, we do not hold the view that there is an underlying true (random) total order together with a coarsening mechanism that generates the (random) partial order. Such views are termed \textit{epistemic view} within the IP-community, cf., \cite{Couso_2014}. Applications of this view in the context of partial order data can be found for example in \cite{NIPS2007_fe8c15fe,nakamura2019learning}. 
Opposed to this view, within the nomenclature of \cite{Couso_2014}, we see our approach more in the spirit of the \textit{ontic view} {that is usually applied to set-valued data and that states that such data are set-valued by nature and that there is no true but unobserved data point within the observed sets. Generally, this distinction between the ontic and the epistemic view is much discussed in the IP comunity and especially at the ISIPTA conferences.\footnote{Concrete applications can be found e.g. in \cite{plass2015_ontic} for the ontic, and in \cite{plass2015statistical} for the epistemic view. A case where the ontic and the epistemic view coincide is discussed in \cite{schollmeyer17a}. Beyond this, in the field of partial identification in the context of generalizing confidence intervals to confidence regions for the so-called identified set, the question about an ontic vs. an epistemic view is also implicitly asked (without reference to these terms) 
by asking if such a confidence set should cover the true parameter or the whole identified set with prespecified probability, cf., \cite{stoye2009statistical}.}} However, since in our case the random objects are partial orders, the term \textit{ontic} as used in \cite{Couso_2014} seems to fit not perfectly. Instead  we understand poset-valued data as a special type of \textit{non-standard data}.\footnote{Note that we do not want to generally rule out epistemic treatment, but this is not the focus of this paper. Such a treatment could use that every partial order can be described by the set of all its linear extensions. For further discussions, c.f.,~\cite{bsj2022}.}

While the same view is taken in~\cite{bsj2022}, the differences here are found more in the objective: while that paper focuses on theory for stochastic modeling of poset-valued random variables, the present paper can be viewed as a framework for analyzing data, e.g., sampled from one of these models.



Of course, a descriptive analysis of samples of partial orders which explicitly addresses the above interpretation requires a completely different -- and so far to the best of our knowledge not existing -- mathematical apparatus compared to an analysis of standard data. A suitable formal framework is by no means obvious here. Fortunately, it turns out that the concept of a \textit{depth function}, which has so far mainly been applied to $\mathbb{R}^d$-valued random variables\footnote{Exceptions are, e.g., the definition of depth functions for functional data in~\cite{lopez09} and on lattices in~\cite{schollmeyer17a, schollmeyer17b}.} (see, e.g.,~\cite{sz2000,m2002}), can be promisingly adapted to poset-valued random variables. Generally speaking, (data) depth functions define a notion of centrality and outlyingness of observations with respect to the entire data cloud.
Equipped with this adapted depth concept, some classical  descriptive statistics can then be naturally adapted to this particular non-standard data type as well. 

Our paper is organized as follows: In Section~\ref{prel}, we briefly discuss the required mathematical definitions and concepts. We give a formal definition of our depth function, the \textit{ufg depth}, in Section~\ref{sec:ufg} and discuss some of its properties in Section~\ref{sec:properties}. The concrete theorems and proofs can be found in the appendix. While Section~\ref{sec:comparison} prepares our application by providing the required background, Section~\ref{sec:application} is devoted to applying our framework to a specific example, namely the analysis of the goodness of classification algorithms on different data sets. Section~\ref{sec:conclusion} concludes by elaborating on some promising perspectives for future research.

\section{Preliminaries} \label{prel}
\textit{Partial orders (posets)} sort the elements of a set $M$, where we allow that two elements $y_1, y_2 \in M$ are incomparable. Formally stated: Let $M$ be a fixed set. Then $p \subseteq M \times M$ defines a partial order (poset) on $M$ if and only if $p$ is \textit{reflexive} (for each $y \in M$ holds $(y, y)\in p$), \textit{antisymmetric} (if $(y_1, y_2), (y_2,y_1) \in P$ then $y_1 = y_2$ is true) and \textit{transitive} (if $(y_1, y_2), (y_2,y_3) \in p$ then also $(y_1,y_3) \in p$). If $p$ is also strongly connected (for all $y_1,y_2 \in M$ either $(y_1,y_2) \in p$ or $(y_2, y_1) \in p$), then $p$ defines a \textit{total/linear order}.
For a fixed set $M$, various posets sort the set $M$. We are interested in all posets that can be obtained for the set $M$, where the cardinality $\# M$ is finite. We denote the set of all posets on $M$ by $\Pcal_M$ (or $\Pcal$ for short). Sometimes it can be useful to consider only the \textit{transitive reduction}, this means that for a poset $p$ we delete all pairs $(y_1, y_2)$ which can be obtained by a transitive composition of two other elements in $p$. Note that there exists a one-to-one correspondence between the transitive reduction of posets and the posets itself. We denote the transitive reduction of a poset $p$ by $tr(p)$. This transitive reduction is often used to simplify the diagram used to represent the partial order. These diagrams are called \textit{Hasse diagram}. They consist of edges and knots where the knots are the elements of $M$ and the edges state which element lies below the other, e.g., see Figure~\ref{fig:max_min}. The reverse, where we add all pairs which follow from transitivity, is called \textit{transitive hull}. We call $th(p)$ the transitive hull of a relation $p$. We refer to \cite{ganter12} for further readings on partial orders.
From now on, let $\Pcal$ be all posets for a fixed set $M$. We denote the elements of $M$ by $y$. 

The analysis concept for poset-valued observations presented here is based on a \textit{closure operator} on $\Pcal$, see Section \ref{sec:ufg}. In general, a closure operator $\gamma_{\Omega}: 2^{\Omega} \to 2^{\Omega}$ on set ${\Omega}$ is an operator which is \textit{extensive} (for $A \subseteq {\Omega}, \: A \subseteq \gamma_{\Omega}(A)$ holds), \textit{increasing}, (for $A,B \subseteq {\Omega}$ with $A\subseteq B,\: \gamma_{\Omega}(A) \subseteq \gamma_{\Omega}(B)$ is true) and \textit{idempotent} (for $A \subseteq {\Omega}, \; \gamma_{\Omega}(A) = \gamma_{\Omega}(\gamma_{\Omega}(A))$ holds). The set $\gamma_{\Omega}(2^{\Omega})$ is called the \textit{closure system}. Note that every closure operator (and therefore the closure system) can be uniquely described by an implicational system. An \textit{implicational system} $\mathcal{I}$ on ${\Omega}$ is a subset of $2^{\Omega} \times 2^{\Omega}$. The implicational system corresponding to the closure operator $\gamma_{\Omega}$ is defined by all pairs $(A,B) \in 2^{\Omega} \times 2^{\Omega}$ satisfying $\gamma_{\Omega}(A) \supseteq \gamma_{\Omega}(B)$. For short, we denote this by $A \to B$. For more details on closure operators and implicational systems, see \cite{bertet18}.

We aim to define a centrality and outlyingness measure on the set of all posets $\Pcal$ based on a fixed and finite set $M$. In general, functions that measure centrality of a point with respect to an entire data cloud or an underlying distribution are called \textit{(data) depth functions}. Depth functions on $\mathbb{R}^d$ have been studied intensively by \cite{sz2000} and \cite{mosler22}, and various notions of depth have been defined, such as Tukeys' depth, see \cite{tukey75}, and simplicial depth, see~\cite{liu90}. The idea behind the ufg depth introduced here is an adaptation of the \textit{simplicial depth} on $\mathbb{R}^d$ to posets, which uses the concept of a closure operator. The simplicial depth on $\mathbb{R}^d$ is based on the convex closure operator which is defined as follows:
	\begin{align*}
		\gamma_{\mathbb{R}^d}\colon \begin{array}{l}
		2^{\mathbb{R}^d} \to 2^{\mathbb{R}^d}\\
		A \mapsto \left\{x \in \mathbb{R}^d \bigg| \begin{array}{l}x = \sum_{i=1}^k \lambda_i a_i \text{ with }  a_i \in A, \\ \lambda_i \in [0,1], \sum_{i=1}^k \lambda_i = 1, k \in \mathbb{N}\end{array} \right\}.
	\end{array}
	\end{align*}
	For the simplicial depth, we consider only input sets $A$ with cardinality $d+1$ which form a $(d+1)$-simplex (when no duplicates occur). Then, the simplicial depth of a point $x \in \mathbb{R}^d$ is the probability that $x$ lies in the codomain of the convex closure operator of $d+1$ points randomly drawn from the underlying (empirical) distribution. The set of all sets $A$ with cardinality $d+1$-simplices is a proper subset of $2^{\mathbb{R}^d}$. By using Carathéodorys' Theorem, see \cite{eckhoff93}, we obtain that any set $B$ of $d+1$ unique points is the smallest set, for which there exists no family of proper subsets $(A_i)_{i \in \{1, \ldots, \ell\}}$ with $A_i \subsetneq B$ such that  $\bigcup_{i \in \{1, \ldots, \ell\}}\gamma_{\mathbb{R}^d}(A_i) = \gamma_{\mathbb{R}^d}(B)$. 
 Thus, these simplices still characterizes the corresponding closure system. For $\mathcal{M}$ being the set of all probability measures on $\mathbb{R}^d$ the simplicial depth is then given by
    		\begin{align*}
    			D\colon  \begin{array}{l} \mathbb{R}^d \times \mathcal{M} \to [0, 1],\\ (x,\nu) \mapsto \nu(x \in \gamma_{\mathbb{R}^d}\{X_1, \ldots, X_{d+1}\}), \end{array}
    		\end{align*}
    		where $X_1, \ldots, X_{d+1} \overset{iid}{\sim} \nu$. When we consider a sample $x_1, \ldots, x_n \in \mathbb{R}^d; \: n \in \mathbb{N}$, we use the empirical probability measure instead of a probability measure $\nu$. Thus, for a sample $x_1, \ldots, x_n \in \mathbb{R}^d$ with empirical measure $\nu_n$ we obtain as empirical simplicial depth
    		\begin{align*}
    			D_n\colon \begin{array}{l}
    			\mathbb{R}^d \to [0,1], \\
    			x \mapsto\binom{n}{d+1} \sum_{1 \le i_1 < \ldots < i_{d+1} \le n} 1_{\gamma_{\mathbb{R}^d}\{x_{i_1}, \ldots, x_{i_{d+1}}\}}(x).
    			\end{array}
    		\end{align*}
    		Hence, if $x_1, \ldots, x_n$ are affine independent, then the depth of a point $x$ is the proportion of $(d+1)$-simplices given by $x_1, \ldots, x_n$ that contain $x$.

\section{Union-Free Generic Depth on Posets} \label{sec:ufg}
Now, we introduce the union-free generic (ufg) depth function for posets which is in the spirit of the simplicial depth function, see Section~\ref{prel}. To define the depth function, we start, similar to the simplicial depth, with defining a closure operator on $\Pcal$. The definition of the closure operator uses formal concept analysis, see \cite{ganter12}, and the formal context introduced in \cite{bsj2022}. This gives us
\begin{align*}
	\gamma\colon \begin{array}{l}
		2^{\Pcal} \to 2^{\Pcal}\\
		P \mapsto \left\{p \in \Pcal \mid \bigcap\limits_{
  \tilde{p}\in P}\tilde{p} \subseteq p \subseteq \bigcup\limits_{\tilde{p} \in P}\tilde{p} \right\}.
	\end{array}	
\end{align*}
This closure operator maps a set of posets $P$ onto the sets of posets where each poset is a superset of the intersection of $P$ and a subset of the union. In other words, any $p$ lying in the closure of $P$ satisfies the following condition: First, every pair $(y_1, y_2) \in M \times M$ that lies in every poset in $P$ is also contained in $p$, and second, for every pair $(y_1, y_2)$ that lies in $p$, there exists at least one $\tilde{p} \in P$ such that $(y_1, y_2)\in\tilde{p}$.
Note that while the intersection of posets defines a poset again, this does not hold for the union. Analogously to the definition of the simplicial depth, we now only consider a subset of $2^{\Pcal}$ and define 
$$\Sscr = \left\{P \subseteq\Pcal \mid \text{Condition } (C1) \text{ and } (C2) \text{ hold } \right\}$$ 
with Conditions (C1) and (C2) given by:
\begin{enumerate}
    \item[(C1)] $P \subsetneq \gamma(P),$
    \item[(C2)] There does not exist a family $(A_i)_{i \in \{1, \ldots, \ell\}}$ such that for all $i \in \{1, \ldots, \ell\}, \: A_i\subsetneq P$ and $\bigcup_{i \in \{1, \ldots, \ell\}} \gamma(A_i) = \gamma(P).$\footnote{In formal concept analysis this is sometimes called \textit{proper}.}
\end{enumerate}
$\Sscr$ is a proper subset of $2^\Pcal$, see Theorem \ref{th: S+card 2 and connect} for details, which reduces $2^\Pcal$ by redundant elements in the following sense: First, all subsets $P\subseteq \Pcal$ with $\gamma(P) = P$ are trivial and therefore not included. Second, if there exists a proper subset $\tilde{P} \subsetneq P$ with $\gamma(\tilde{P}) = \gamma(P)$, then $P$ is also not in $\Sscr$. This follows by setting $\ell =1$ and $A_1 = \tilde{P}$, which defines a family contradicting Condition (C2). These two properties can be generalized to arbitrary closure systems, and referring to \cite{bastide00}, we call a set fulfilling these properties \textit{generic}. The final reduction is to delete also all sets $P$ where $P$ can be decomposed by a family of proper subsets $(A_i)_{i \in \{1, \ldots, \ell\}}$ of $P$. Further, in this case, the union of $(\gamma(A_i))_{i \in \{1, \ldots, \ell\}}$ equals $\gamma(P)$. Note that due to extensitivity, the assumption $\cup_{i \in \{1, \ldots, \ell\}}\gamma(A_i) \subseteq \gamma(P)$ is always true. We call sets respecting this third part \textit{union-free}. Thus $\Sscr$ consists of elements which are \textit{union-free and generic}.

\begin{example}\label{example}
    As a concrete example, consider the set $\Sscr$ based on all posets on $\{y_1, y_2, y_3\}$. Let $p_1, p_2$ and $p_3$ be posets given by the transitive hull of $\{(y_1, y_2)\}$, $\{(y_1, y_2), (y_1, y_3)\}$ and $\{(y_1, y_3), (y_2, y_3)\}$. One can show that the closure of the family $\{p_1, p_3\}$ gives the same closure as $\{p_1, p_2, p_3\}$. Thus, $\{p_1, p_2, p_3\}$ contradicts Condition (C2). For a single poset $p$ we can immediately prove that the closure contains only itself. Therefore, any set consisting of only one poset does not satisfy Condition (C1). In contrast, $\{p_2, p_3\}$ satisfies both Condition (C1) and Condition (C2), since it implies the trivial poset $p_\Delta:=\{(y,y)\mid y \in M\}$, consisting only of the reflexive part. Thus, $\{p_1, p_2\}$ is an element of $\Sscr$.
\end{example}


Now, we define the \textit{union-free generic (ufg) depth} of a poset $p$ to be the weighted probability that $p$ lies in a randomly drawn element of $\Sscr$. Let $\mathcal{M}$ be the set of probabilities on $\Pcal$. The \textit{union-free generic (ufg for short) depth on posets} is given by 
\begin{align*}
	D\colon \begin{array}{l}
	 \Pcal \times \mathcal{M} \to [0,1] \\
	 (p, \nu) \mapsto \begin{cases}
            0, \quad \text{if for all } S\in \Sscr\colon \prod_{\tilde{p} \in S}\nu(\tilde{p}) = 0\\
	      c\sum_{S \in \Sscr \colon p \in \gamma(S)} \prod_{\tilde{p} \in S}\nu\left(\tilde{p}\right),\quad \quad \text{else,} 
  \end{cases}
\end{array}		
\end{align*}
 with $c = \left(\sum_{S \in \Sscr} \prod_{\tilde{p} \in S}\nu_n\left(\tilde{p}\right)\right)^{-1}$\footnote{Note that Condition (C1) and (C2) can be applied to the convex closure operator on $\mathbb{R}^d$, see Section \ref{prel}, and we obtain an adapted $\Sscr_{convex}$. Then, $\Sscr_{convex}$ together with  $\mathcal{M}_{convex}$ the set of measures which are absolute continuous to the Lebesgue measure, leads to the simplicial depth.}. These two cases are needed because $c$ is not defined in the first case. Note that if there exists an $S \in \Sscr$ with $\prod_{\tilde{p} \in \Sscr}\nu(\tilde{p}) \neq 0$, then $D \not\equiv 0$. The case that $D \equiv 0$ only occurs in two specific situations which result from the structure of the probability mass, see Property (P2) and Corollary~\ref{cor: Dn always zero} for details. In contrast to the simplicial depth where only sets of cardinality $d+1$ are considered, the elements of $\Sscr$ differ in their cardinality. Thus, different approaches on how to include the different cardinalities are possible, i.e., by weighting. Here, we used weights equal to one.

The empirical version of the ufg depth uses the empirical probability measure $\nu_n$ given by an iid~sample of posets $\underline{p} = (p_1, \ldots, p_n), \: n \in \mathbb{N}$ instead. We obtain as \textit{empirical union-free generic (ufg) depth}
\begin{align*}
	D_n \colon \begin{array}{l}
		\Pcal \to [0,1] \\
		p \mapsto  \begin{cases}
            0, \quad \text{if for all } S\in \Sscr\colon \prod_{\tilde{p} \in S}\nu_n(\tilde{p}) = 0\\
            c_n \sum_{S \in \Sscr, p \in \gamma(S)} \prod_{\tilde{p} \in S}\nu_n\left(\tilde{p}\right), \qquad \quad \text{else,}
              \end{cases}
	\end{array}
\end{align*}
 with $c_n = \left(\sum_{S \in \Sscr} \prod_{\tilde{p} \in S}\nu_n\left(\tilde{p}\right)\right)^{-1}$. The empirical ufg depth of a poset $p$ is therefore the normalized weighted sum of drawn sets $S \in \Sscr$ which imply $p$. Note that when restricting $\Sscr$ to the set $\{S \cap \{p_1, \ldots, p_n\} \mid S \in \Sscr\}$, this does not change the depth value. This holds since for other elements $S \in \Sscr$, the empirical measure for at least one $p \in S$ is zero.
 
\begin{example}\label{example_2}
    Returning to Example~\ref{example}, suppose that we observe $(p_1, p_2, p_3)$. Then for the trivial poset $p_\Delta$, the empirical depth is $D_n(p_\Delta) = 1/2$. For the set $p_4$ given by the transitive hull of $\{(y_3, y_1)\}$, the value of the empirical depth is zero.
    For $p_{total}$ given by the transitive hull of $\{(y_1, y_3), (y_3, y_2)\}$, the empirical depth value is again zero.
\end{example}

\section{Properties of the UFG Depth and $\Sscr$}\label{sec:properties}
For a better understanding of the ufg depth, we now discuss some properties of $D_n$ and $\Sscr$. The properties of $D_n$ describe the mutual influence between the (empirical) measure and the ufg depth while the properties of $\Sscr$ can be used to improve the computation time.

\subsection{Properties of the (Empirical) UFG Depth}
The following statements are given for $D_n$. Those properties which focus on the empirical measure and not on the concrete sample values can be transferred to $D$.
The first observation is that the ufg depth 
\begin{center}
    (\textbf{P1}) \textit{considers the orders as a whole, not just pairwise comparisons.}
\end{center}
More precisely, the ufg depth cannot be represented as a function of the sum-statistics $$\left(w_{(a,b)} := \# \{i \in\{1,\ldots,n \} \mid (a,b) \in p_i\}\right)_{(a,b) \in M \times M}$$ of the pairwise comparisons,\footnote{Note that many classical approaches rely only on the sum-statistics. For example within the Bradly-Terry-Luce model (cf.,~\cite[p. 325]{btl}) or the Mallows $\Phi$ model (cf.,~\cite[p. 360]{fligner}), the likelihood function that is maximized depends only on the data through the sum-statistics.}  see Theorem~\ref{thm_pairwise}.

Remarkably, this concretization of this property formalizes precisely the analogy to the ontic notion of non-standard data mentioned at the beginning: Computing the depth of a partial order cannot be broken down via simple sum-statistics, but requires the partial order as a holistic entity. This is due to the fact that the involved set operations within the closure operator $\gamma$ rely on the partial orders as a whole.

In Section \ref{sec:ufg}, we defined the ufg depth in terms of two cases. If there exists at least one element $S \in \Sscr$ such that every $p \in S$ has a positive empirical measure, then $D_n \not\equiv 0$. In Corollary~\ref{cor: Dn always zero} we specify this
\begin{center}
    (\textbf{P2}) \textit{non-triviality property}.
\end{center}
We claim that $D_n \equiv 0$ occurs only when either the entire (empirical) probability mass lies on one poset or when the (empirical) probability mass is on two posets where the transitive reduction differ only in one pair.

The next observation relates to how the sampled posets affect the ufg depth value. For example, let us recall Example~\ref{example} and Example~\ref{example_2}. From the structure of the sample, we can immediately see that $p_\Delta$ has a nonzero depth and that $p_{total}$ must have a depth of zero. For this
\begin{center}
    \textbf{(P3)} \textit{implications of the sample on }$D_n$ \textit{property,}
\end{center}
let $\underline{p}= (p_1, \ldots, p_n)$ be a sample from $\Pcal$. Let $(y_1,y_2) \in M\times M$ such that for all $i \in \{1, \ldots, n\}, \: (y_1,y_2) \not\in p_i$. Then for every $p\in \Pcal$ with $(y_1,y_2) \in p$, we get $D_n(p) = 0$. This means that if a pair does not occur in any poset of the sample, then every poset which contains this pair needs to have zero empirical depth.
Reverse, when looking at non-pairs, a similar statement is true. Let $p \in \Pcal$ with $(y_1, y_2) \not\in p$ but for all $i \in \{1, \ldots, n\}$, $(y_1, y_2) \in p_i$ holds. Then, $D_n(p) = 0$.
This follows from Corollary \ref{cor: Dn and the sample}.
The influence of duplicates on the value of the empirical ufg depth ${D_n}$ is immediately apparent by using the empirical measure $\nu_n$. Thus, each element in $\Sscr$ is weighted by the number of duplicates in the sample $\{p_1, \ldots, p_n\}$.

Conversely to Property (P3), in some cases, structure in the sample can be inferred by the ufg depth values. In Example~\ref{example} and Example~\ref{example_2}, knowing only the values of the depth function gives us some insight into the observed posets. For example, we know that there must be at least one pair $(y_i, y_j)$ that is an element of $p_{total},$ but which is not given by any observed poset. Moreover, the fact that $p_\Delta$ has nonzero depth implies that there exists no pair $(y_i, y_j)$ that every observed poset has. We call this property
\begin{center}
    \textbf{(P4)} \textit{implications of the outliers on the sample.}
\end{center}
More precisely, the depth value of the trivial poset, which consists only of the reflexive part, as well as the values of the total orders, can provide further information about the sample. Therefore, let $p_{\Delta}$ be the trivial poset, and $p_{\text{total}}$ be a total order. By Corollary \ref{cor: Dn and the sample} we obtain that if $D_n(p_{\Delta}) = 0$, then there exists at least one pair $(y_1, y_2)$ which is in every poset of the sample. The knowledge of $p_{total}$ leads to an statement about the non-edges. So, if $D_n(p_{\text{total}}) = 0$ is true, then there exists at least one pair $(y_1, y_2) \in p_{\text{total}}$ which is in no poset of the sample.

The last properties have summarized how the structure of a sample is reflected in the ufg depth and vice versa. Finally, we have
\begin{center}
    \textbf{(P5)} \textit{consistency of the empirical ufg depth $D_n$.}
\end{center} 
This means that $D_n$ converges uniformly to $D$ almost surely under the assumption of observing i.i.d. samples, see Theorem~\ref{th: convergence}.

\subsection{Properties of $\Sscr$}
In this subsection, we introduce some properties of $\Sscr$, which we use to improve the computation. The first one is 
\begin{center}
    \textbf{(P6)} \textit{a lower bound for all} $S \in\Sscr ,$ 
\end{center}
which is given by ${\#S\ge 2}$. This fact is already discussed in Example~\ref{example}.
For the upper bound we use a complexity measure of $\Sscr$, the Vapnik-Chervonenkis dimension (VC dimension for short), see \cite{Vapnik15}. The VC dimension of a family of sets $\Cscr$ is the largest cardinality of a set $A$, such that $A$ can still be shattered into the power set of $A$ by $\Cscr$.\footnote{To be more precise: The intersection between a set $A$ and a family of sets $\Cscr$ is defined by $A \cap \mathcal{C} = \{A \cap C \mid C \in \Cscr\}$. We say that a set $A$ can be shattered (by $\Cscr$) if $\# (A \cap \Cscr) = 2^{\# A}$ holds. The VC dimension of $\Cscr$ is now defined as $vc = \max \{\# A\mid (A \cap \Cscr) = 2^{\# A}\}.$}
With this, we obtain
\begin{center}
    \textbf{(P7)} \textit{an upper bound for all} $S \in \Sscr $ 
\end{center}
 is given by ${\#S \le vc}$, with $vc$ the VC dimension of the closure system $\gamma(2^{\Pcal})$.
The proof of the upper and lower bound can be found in Theorem \ref{th: upper, lower bounds S}. Note that in our case of posets, the VC dimension is small compared to the number of all posets.

 We conclude with the observation that
 \begin{center}
     \textbf{(P8)} \textit{the elements of} ${\Sscr}$ \textit{are connected}
 \end{center}
in the sense that for every $S \in \Sscr$ with $\#S = m \ge 3$, there is an $\Tilde{S} \in \Sscr$ such that $\Tilde{S} \subsetneq S$ and $\#\Tilde{S} = m-1$, see Theorem \ref{th: S+card 2 and connect}.

\section{Comparing Machine Learning Algorithms} \label{sec:comparison}
Before turning to our actual application, we first indicate, which possible contributions our methodology based on data depth in the context of poset-valued data is able to add to the general task of analyzing machine learning (ML) algorithms beyond pure benchmarking considerations. The basic task of performance comparison of algorithms is very common in machine learning (cf., \cite{Hothorn} and the references therein). Our methodological contribution deviates from the typical benchmark setting with regard to at least two points:

(I) First, we compare algorithms not with respect to one unidimensional criterion like, e.g., balanced accuracy, but instead we look at a whole set of performance measures. 
We then judge one algorithm as at least as good as another one if it is not outperformed with respect to any of these performance measures. With this, for every data set, we get a partial order of algorithms and since we
are not looking at 
only one, but a whole population/sample of data sets, we get a poset-valued random variable.

(II) Second, we are not interested in the question which algorithm is in some certain sense the best or competitive, etc. Instead, we are interested in the question how the relative performance of different algorithms is distributed over a population/sample of different data sets. Analyzing the distribution of performance relations is in our view a research question of its own statistical importance that may add further insights to analyses in the spirit of e.g., \cite{jnsa2022} which are of most importance when it comes to choosing between different machine learning algorithms.\\

\noindent These both deviations can have very different motivations: The analysis of a multidimensional criteria (of performance, here) is already motivated in the fact that in a general analysis, different performance measures are in the first place conceptually on an equal footing (at least, if one has no further concrete, e.g., decision-theoretic desiderata at hand). Therefore it appears natural to take more than one measure at the same time into account. Beyond this, there are far more 
possible motivations for dealing with multidimensional criteria of performance:
For example for classification, if one accounts for the impact of distributional shifts within covariates, then one aspect to consider is that for different covariate distributions, the class balance of the class labels will vary, which can naturally be captured by either looking at different weightings of the true positives and the true negatives within the construction of a classical performance measure\footnote{Note that usual performance measures are more or less simple transformations of the vector of the true positives and the true negatives (and the class balance).} or alternatively by taking into account different discrimination thresholds for the classifiers simultaneously, which would correspond to looking at a whole region of the receiver operating characteristic. 

Also the motivation for the second point can be manifold:
Generally, it seems somehow naive to search for one best algorithm per se. For example, the scope of application of an  algorithm can vary very strongly and therefore, for different situations, different algorithms could be the best, or in certain situations different algorithms can be comparable in its performance, or, on the other hand, incomparable if one looks at different performance measures at the same time. Generally, it can be of high interest, how the conditions between different algorithms change over the distribution of different data sets or application scenarios. For example, if in one very narrowly described data situation the performances of different algorithms vary extremely from case to case, but not so much across algorithms, then, at some point it would become more or less hopeless to search for a best algorithm in the training phase, because one knows that in the prediction setting, the situation is too different to the training situation.

Another aspect is outlier detection: If one knows that in a large, maybe automatically generated benchmark suite there are data sets that have some bad data quality (for example if some covariates are meaningless because of some data formatting error etc.), then, it would be reasonable to try to exclude such outlying data sets from a benchmark analysis beforehand.  Candidates of such outliers are then naturally data sets with a low depth value. 


\section{Application on Classifier Comparison} \label{sec:application}
After this motivation, we now apply our ufg depth on poset-valued data: each poset arises from comparison of classifiers based on multiple performance measures on a data set. 


\subsection{Implementation}\label{subsec:implementation}
Let $(p_1, \ldots, p_n)$ be a sample of posets.
There are two difficulties in computing $D_n$. First, going through each subset of $ \{p_1, \ldots, p_n\}$ is very time-consuming, especially since the subsets that are an element of $\Sscr$ can be very sparse in $2^{\{p_1, \ldots, p_n\}}$. Second, it is difficult to test whether a subset is an element of $\Sscr$ or if it is not an element. 

The first part can be improved by using the lower and upper bound on the cardinality of $S \in \Sscr$, see Section \ref{sec:properties}. Here we use a binary integer linear programming formulation described in \cite[p.33f]{epub40416} to compute the VC dimension. Further, we use the connectedness of the elements $S \in \Sscr$, see Property (P8) in Section \ref{sec:properties}. With this, we do not have to go through every subset that lies between the lower and upper bounds, but can stop the search earlier.

To check whether a subset $P \subseteq \{p_{1} \ldots, p_{n}\}$ is an element of $\Sscr$, we begin with two observations: First, Condition (C2) implies that there must exist a poset $p$ that does not lie in any closure operator output of any proper subset of $P$. (This follows from the extensivity of the closure operator~$\gamma$.) Second, Condition (C2) implies Condition (C1) for $\ell \ge 2$, since only for $\ell = 1$ we cannot define a family $(A_i)_{i \in \{1, \ldots, \ell\}}$ consisting of proper subsets of $P$ such that for every $p \in P$ there exists an $i \in \{1, \ldots, n\}$ with $p \in A_i$.

By Property (P6), we know that for any $S \in \Sscr$, $\# S \ge 2$ is true. So we only need to check if Condition (C2) is true. Thus, we want to find a poset $p$ which is given only by the entire set $P.$ Suppose that $p$ is such a poset. Then $\bigcap\limits_{\tilde{p} \in P} \tilde{p} \subseteq p \subseteq \bigcup\limits_{\tilde{p} \in P} \tilde{p} $ and for every $\tilde{p} \in P$ at least one of the following statements is true:
\begin{enumerate}
    \item[$(T1)_{\tilde{p}}$] There exists a pair $(y_1, y_2) \in \tilde{p}$ with $(y_1, y_2) \in p$. But for all other $\hat{p} \in P\setminus\tilde{p}$, $(y_1, y_2) \not\in \hat{p}$ is true. 
    \item[$(T2)_{\tilde{p}}$] There exists a pair $(y_1, y_2) \not\in \tilde{p}$ with $(y_1, y_2) \notin p$. But for all other $\hat{p} \in P\setminus\tilde{p}$, $(y_1, y_2) \in \hat{p}$ is true. 
\end{enumerate}
Thus, none of these $\tilde{p}$'s can be deleted since then $P \setminus \tilde{p}$ does not contain $p$ in the closure output of $\gamma$ anymore. 
After this analysis of the candidate $p$. we are now interested in the construction of the candidate $p$. First, observe that for every $\tilde{p} \in \Pcal$ and $i \in \{1,2\}$ one can collect all pairs $(y^{(Ti)_{\tilde{p}}}_1,y^{(Ti)_{\tilde{p}}}_2)$ which can be used to ensure that $(Ti)_{\tilde{p}}$ holds. Set $p_M = \cap_{\tilde{p} \in P}\:\tilde{p}$. For every element $\tilde{p} \in P$ choose one pair $(y^{(Ti)_{\tilde{p}}}_1,y^{(Ti)_{\tilde{p}}}_2)$. Now, add those pairs to $p_M$ with $i = 1$. Compute $th(p_M)$ and for every $\tilde{p}$ for which we chose $(y^{(Ti)_{\tilde{p}}}_1,y^{(Ti)_{\tilde{p}}}_2)$ with $i = 2$, check that $\tilde{p}$ is necessary to obtain $p_M$ in the output of the closure operator.
If this holds for any $\tilde{p}$ where a pair with $i = 2$ is chosen, and $th(p_M) \subseteq \cup_{\tilde{p} \in P} \: \tilde{p}$, then $p_M$ is a poset that ensures Condition (C2). Thus, by going through all combinations of $(y^{(Ti)_{\tilde{p}}}_1,y^{(Ti)_{\tilde{p}}}_2)$, we can check whether such a poset exists. It is sufficient to pick for each $\tilde{p}$ precisely one $(y_{\tilde{p}_1}, y_{\tilde{p}_2})$ since $th(\hat{p}) \subseteq th(\tilde{p})$ is true if $\hat{p}\subseteq \tilde{p}$.

All in all, we improved the computation compared to the naive approach by using the knowledge provided in Section~\ref{sec:properties}. 
Now, we can specify a worst and best case by the bounds. By further including the improved testing of Condition (C1) and (C2) and the connectedness property, we could decrease the computation time, although we currently cannot calculate the exact amount in general as this depends on the complexity of the data set used. Note that even the upper bound is not fixed, but depends on the structure of the data set.
In our application, see Section~\ref{sec:dataset} and \ref{sec:analysis}, we consider 80 posets. The naive approach is not computable in reasonable time since one would have to compute and test each subset of the 80 posets. 
The approach above then leads to a computation time of approximately four hours.

\subsection{Data Set}\label{sec:dataset}
To showcase the application of the ufg depth on machine learning algorithms we use openly available data from the OpenML benchmarking suite \cite{OpenML2013}.

In our comparison we include the following supervised learning methods: \textit{Random Forests} (RF, implemented in the R-package \texttt{ranger} \cite{ranger}), \textit{Decision Tree} (CART, implemented via the \texttt{rpart} library \cite{therneau2015package}),  \textit{Logistic regression} (LR), \textit{L1-penalized logistic regression} (Lasso, implemented through the \texttt{glmnet} library \cite{friedman2021package}) and \textit{k-nearest neighbors} (KNN, through the \texttt{kknn} library \cite{epub1769}).
As stated in the OpenML experiment-documentation all methods are run with default settings of the corresponding libraries. Hence, our application analyzes the behavior of methods using default settings and does not necessarily extend to general statements about the performance of hyperparameter-tuned versions of the respective algorithms. The algorithms were chosen as a selection of widely used supervised learning methods that perform reasonably without much tuning, in contrast to methods such as neural networks or boosting, which require considerable tuning to perform well.

From the available data sets for which results for all above algorithms are available in the OpenML database, we limit our analysis to binary classification data sets with more than 450 and less than 10000 observations, leading us to a total of 80 data sets for comparison. The data sets come from a variety of domains and strongly vary in their class balance as well as their overall difficulty.
Included in our multidimensional criteria comparison are the measures \textit{area under the curve}, \textit{F-score}, \textit{predictive accuracy} and \textit{Brier score}. These performance measures capture different aspects of performance, especially in the case of unbalanced data sets.

The corresponding posets then result from the multidimensional criteria comparison. It should be noted that rescaling the performance measures does not change the posets. This follows from the fact that the posets defined here do not depend on the absolute differences but on whether or not the multiple performance measures are better in all dimensions.

\subsection{Analysis}\label{sec:analysis}
\begin{figure}
    \centering
    \includegraphics[scale = 0.6]{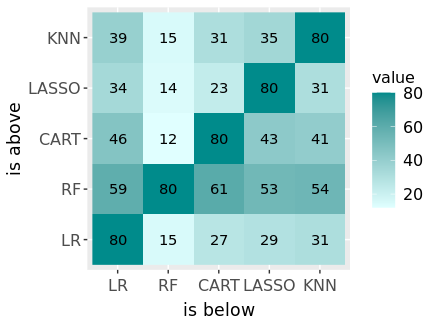}
    \caption{Heatmap representing the sum-statistics, see Footnote~$9$.}
    \label{fig:heatmap}
\end{figure}

The resulting poset-valued set consists of $80$ posets, $58$ of which are unique. Each of the $58$ unique posets have a different depth value. The sum-statistics, see Footnote~$4$, which count for each pair the number of occurrences along the $80$ posets, can be seen in Figure~\ref{fig:heatmap}. It shows that RF is very often above all other methods. So if one only looks at the sum-statistics RF is clearly the strongest method. The other methods are more balanced with respect to each other. Note that due to reflexivity the diagonal is always $80$.

\begin{figure}
    \centering
    \includegraphics[scale = 0.6]{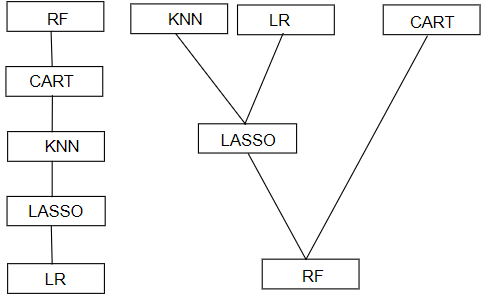}
    \caption{Observed poset with maximal (left) and minimal (right) ufg depth. On the left-hand side RF dominates every other algorithm and in contrast on the right-hand side RF is dominated by every other.}
    \label{fig:max_min}
\end{figure}

The most central poset with the maximum depth value is a total order and can be seen in Figure~\ref{fig:max_min}. Its depth value is 0.34. 
The poset with the highest depth value also has the most duplicates, meaning it is the most common pattern. As described in Section~\ref{sec:comparison}, we are interested in the distribution of the observed posets. Nevertheless, we can consider the poset with the highest depth value as the poset whose structure is the most common one. Or, in other words, this poset is the one that is most supported based on all observations. Comparing this to Figure~\ref{fig:heatmap} or, e.g., the results in~\cite{jnsa2022}, we see many similarities, such as LR often has worse performance than the other algorithms, and RF dominates all other algorithms in many cases. 
In contrast to the sum-statistics which here give a representative poset, the strength of our method is that we not only obtain one single poset structure, but also a distribution over the set of posets. Note that in general the order given by the sum-statistics is not a poset, i.e. their might exist cycles.

\begin{figure}
    \centering
    \includegraphics[scale = 0.5]{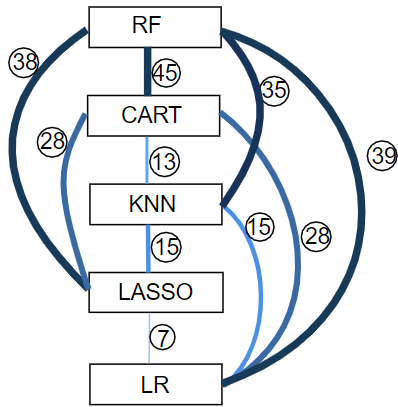}
    \caption{Represents what the observed posets with the $k$ highest depth values have in common. Compare with Figure~\ref{fig:max_min}, where the poset with the highest depth value is plotted. Here each edge number $k$th indicates that the k deepest posets all contain this relation, but this is not true for the $k+1$ deepest poset.}
    \label{fig:Schnitt}
\end{figure}

Figure \ref{fig:Schnitt} describes which edges the posets with the $k \in \{1, \ldots, 80\}$ highest depth values have in common. For example, one can observe that the dominance of RF over all classifiers based on all four performance measures holds for the 35 posets with the highest depth values. In particular, any other classifier dominance (like CART outperforms KNN according to all performance measures), does not hold for the 35 posets with the highest depth values. For example that CART outperforms KNN is only true for the 13 deepest posets. Note that the posets with the highest 46 depth values have nothing more in common. Conversely, it is of interest to see what non-edges the posets have in common. Since the poset with the highest depth value is the total order, this is immediately apparent in Figure~\ref{fig:Schnitt}. The posets with $k \in \{1, \ldots, 80\}$ highest depth values have those non-edges in common, which are given by the inversely ordered poset of highest depth value intersecting with the inversely already deleted ones. For example, the posets with the nine highest depth values have in common that the RF is not dominated by CART, CART not by KNN and KNN not by Lasso, but they do not agree on LASSO being not dominated by LR since the poset with the 8th highest depth value does not agree on this.

Unlike the posets with the highest depth values, the posets with low depth values do not have much in common. The posets with the tenth lowest depth values only agree on RF being dominated by another classifier. After that, no structure holds. All of these posets can be seen as outlier, or in other words, the corresponding data sets produce a performance structure on the classifiers which differ from the structure given by other data sets. The poset with the smallest depth value, which is 0.05, can be seen on the right side of Figure~\ref{fig:max_min}. 

Finally, we want to give a notion of dispersion of the depth function. Therefore, we compute the depth function for every poset $p \in \Pcal$ and compute the proportion of posets which lie in $\alpha \in [0,1]$ deepest observed depth values. For $\alpha = 0.25, 0.5$ and $0.75$ we get $0.02, 0.10$ and $0.26$. Thus, the empirical ufg depth seems to be clustered on small parts of the set $\Pcal$.

 To summarize, the concept of depth functions allowed us to get valuable insight in the typical order of the analyzed classifiers. Further, it detects data sets where the structure of the classifiers, given by the performance measures, seems to have unusual structure.

\section{Conclusion} \label{sec:conclusion}



In this paper, we have shown how samples of poset-valued random variables can be analyzed (descriptively) by utilizing a generalized concept of data depth. For this purpose, we first introduced an adaptation of the simplicial depth, the so-called ufg depth, and studied some of its properties. Finally, we illustrated our framework with the example of comparing classifiers using multiple performance measures simultaneously. There are several promising avenues for future research, that include (but are not limited to):

\textbf{Other ML Problems and Criteria:} Here, we focused on the comparison of classifiers by a set of unidimensional performance criteria. For example, the performance of different optimization algorithms could be also of interest. Further, the analysis of classifiers with respect to other criteria could be an interesting modification. For example one could use ROC curves or criteria that do also take the fact into account that classical performance measures are only estimates of the true out of sample performance. Within our order-based approach this would be easily incooperateable.

\textbf{Discussion on computation time:} In Section~\ref{subsec:implementation} we briefly discussed the computation time and the difficulty of predicting it. For a deeper understanding further analyses, e.g. in form of a simulation study, would be helpful.

\textbf{Inference:} A first step towards inference for poset-valued random variables is already made by the consistency property in Section~\ref{sec:properties}. Natural next tie-in points are provided by regression and statistical testing. Together with the results for modeling in~\cite{bsj2022}, a complete statistical analysis framework for poset-valued random variables would then be achieved.

\textbf{Other types of non-standard data:} Our analysis framework is by no means limited to poset-valued random variables. Since the ufg depth is based on a closure operator, all non-standard data types for which a meaningful closure operator exists can be analyzed with it. As seen in \cite{bsj2022} such closure operators are easily obtained by formal concept analysis, thus, there exists a natural generalization of the ufg depth for non-standard data.

\appendix

\section{Proofs of Section \ref{sec:properties}}
The next part presents the proof of the properties given in Section~\ref{sec:properties}. First, for a fixed $p \in \Pcal$, we give a different representation of the sets $S \in \Sscr$ with $p \in S$.
\begin{lemma}\label{th: restructuring Sscr}
	For $p \in \Pcal$ we get
        \begin{align}\label{eq:Sscr als Schnitt_1}
	&\{S\in \Sscr\mid p \in \gamma(S)\} \\
	&= \bigcap\limits_{(y_i, y_j) \in p}\{S\in \Sscr\mid \exists x \in S\colon (y_i, y_j) \in x\}\: \cap \label{eq:Sscr als Schnitt_2}\\
	&\quad \bigcap\limits_{(y_i, y_j) \not\in p}\{S\in \Sscr\mid \exists x \in S \colon (y_i, y_j) \not\in x\} \label{eq:Sscr als Schnitt_3}.
\end{align}
\end{lemma}
\begin{proof}
Let $p \in \Pcal$. The proof is divided into two parts. \\
Part 1: We prove $\subseteq$. Let $S$ be an element of (\ref{eq:Sscr als Schnitt_1}). Since $p \in \gamma(S)$, we have $p \subseteq \cup_{\tilde{p} \in S} \:\tilde{p}.$ So for every $(y_i,y_j) \in p$ there is a $\tilde{p} \in S$ such that $(y_i, y_j) \in \tilde{p}$. Therefore, $S$ is an element of the intersection of (\ref{eq:Sscr als Schnitt_2}). Also from $p \in \gamma(S)$ we get $\cap_{\tilde{p} \in S}\:\tilde{p} \subseteq p$ and thus we know that for every $(y_i,y_j) \not\in p$ there exists a $\tilde{p} \in S$ such that $(y_i, y_j) \not\in \tilde{p}$. Thus, $S$ is an element of the intersection given by (\ref{eq:Sscr als Schnitt_3}).\\
Part 2: We prove $\supseteq$. Therefore, let $S\in \Sscr$ be an element of the right-hand side of the equation. We show that $p \in \gamma(S)$. Let $S$ be in the intersection given by (\ref{eq:Sscr als Schnitt_2}). Then we know that for every $(y_1,y_2) \in p$ there exists an $\tilde{p}\in S$ such that $(y_1, y_2) \in \tilde{p}$. Thus $p \subseteq \cup_{\tilde{p} \in S}\: \tilde{p}$. The second part of the intersection given by (\ref{eq:Sscr als Schnitt_3}) analogously yields that $\cap_{\tilde{p}\in S} \: \tilde{p} \subseteq p$. Hence $p \in \gamma(S)$ and the second part is proven. The claim follows from Part 1 and Part 2.
\end{proof}

The next theorem provides some information about the properties of the elements in $\Sscr$. 
\begin{theorem}\label{th: S+card 2 and connect}
The family of sets $\Sscr$ given in Section \ref{sec:ufg} fulfills the following properties.
    \begin{enumerate}
        \item For every $p \in \Pcal$, $\{p\} \not\in \Sscr$.
        \item Let $\{p_1, p_2\} = S\in 2^\Pcal$. Then $S \not\in \Sscr$ iff
        the transitive reductions $tr(p_1)$ and $tr(p_2)$ differ only on one $(y_i,y_j)$ which is only contained in exactly either $tr(p_1)$ or $tr(p_2)$. This means that either $\#(tr(p_1)\setminus tr(p_2)) = 1$ or $\#(tr(p_2) \setminus tr(p_1)) = 1$ holds.
        \item $\mathscr{S}$ is connected in the sense that for every set $S\in\mathscr{S}$ of size $m\geq 3$ there exists a subset $\Tilde{S}\subsetneq S$ of size $m-1$ that is in $\mathscr{S}$, too.
    \end{enumerate}
\end{theorem}
\begin{proof}
    Claim 1. follows directly from Condition (C1) of the definition of $\Sscr$ as $\gamma(\{p\}) = \{p\}$ for every $p \in \Pcal$.

Now, we show the second claim. Let us first assume that $\{p_1, p_2\} = S \not\in \Sscr$. Then there exists no $p \in \Pcal$ such that $p \in \gamma(S) \setminus\{p_1, p_2\}$. Thus, the intersection must be either $p_1$ or $p_2$, (otherwise $p_1\cap p_2 \in \gamma(S)\setminus S$). W.l.o.g., let $p_1 = p_1 \cap p_2$. Then $p_2$ must be a superset of $p_1$ where there is no poset lying between $p_1$ and $p_2$. Therefore, $\#\{tr(p_2)\setminus tr(p_1)\} = 1$ is true.
    Conversely, assume that $S \in \Sscr$ and that $p_1$ is a superset of $p_2$. With this we obtain that $\gamma(S) = \{p \in \Pcal\mid p_2 \subseteq p \subseteq p_1\}$. Further assume that $\# \{tr(p_1)\setminus tr(p_2) \}= 1$ holds. But then $\gamma(S) = S$ is true, since no $p \in \Pcal$ can lie between $p_1$ and $p_2$, but this is a contradiction which proves the claim.

    The proof of the last part uses that the closure operator $\gamma$ stems from a formal context, which is a term from formal concept analysis. Since formal concept analysis is not part of this paper, we have outsourced the proof to \cite{Blocher_note}.
\end{proof}

Using the theorem above, we can determine properties for $\nu$ such that $D \equiv 0$ is true.
\begin{corollary}\label{cor: Dn always zero}
   $D(p) = 0$ for every $p \in \Pcal$ iff the measure $\nu$ has either the entire positive probability mass on a single poset $p$ or only on exactly two posets $p_1$ and $p_2$ where the transitive reduction differs only in a pair $(y_1,y_2)$. More precisely, either $\#\{tr(p_1) \setminus tr(p_2)\} = 1$ or $\#\{ tr(p_2)\setminus tr(p_1)\} = 1$.
\end{corollary}
\begin{proof}
    Note that $D(p) = 0$ for every $p \in \Pcal$ is true if for all $S \in \Sscr$, $\prod_{\tilde{p} \in S}\nu(\tilde{p}) = 0$. Theorem \ref{th: S+card 2 and connect} 1. and 2. provide the cases when this holds which proves immediately the claim.

    The converse follows analogously from Theorem \ref{th: S+card 2 and connect}.
\end{proof}

We use Lemma~\ref{th: restructuring Sscr} to prove the sample influence:
\begin{corollary}\label{cor: Dn and the sample}
	Let $(p_1, \ldots, p_n)$ be a sample of $\Pcal$. Let $\nu_n$ be the empirical probability measure induced by $(p_1, \ldots, p_n)$. Furthermore, let $\nu_n$ be such a probability measure that $D_n \not\equiv 0$. Then for $D_n$ defined in Section \ref{sec:ufg}, it holds.
	\begin{enumerate}
		\item Assume that for all $p_i \in \{p_1, \ldots, p_n\}$, $(y_1, y_2) \in p_i$ is true. Then for every poset $p \in \Pcal$ with $(y_1, y_2) \not\in p$, $D_n(p) = 0$ follows.
		\item Assume that for all $p_i \in \{p_1, \ldots, p_n\}$, $(y_1, y_2) \not\in p_i$ holds. Then for every poset $p \in \Pcal$ with $(y_1, y_2) \in p$, $D_n(p) = 0$ is true.
		\item  Let $p_{\Delta}$ be the poset consisting only of the reflexive part. If $D_n(p_{\Delta}) = 0$, then there exists a pair $(y_1, y_2)$ such that for all $p_i \in \{p_1, \ldots, p_n\}, (y_1, y_2) \in p_i.$
		\item Let $p_{total} \in \Pcal$ be a total order. If $D_n(p_{total}) = 0$, then there exists a pair $(y_1, y_2)\not\in p_{total}$ such that for all $p_i \in \{p_1, \ldots, p_n\}, (y_1, y_2) \in p_i$ is true.
	\end{enumerate}
\end{corollary}
\begin{proof}
    First, note that for $S \in \Sscr$, where there exists an $\tilde{p} \in S$ such that $\nu_n(\tilde{p}) = 0$, $S$ contributes nothing to $D_n$. So one can replace $\Sscr$ in the definition of $D_n$ by $\Tilde{\Sscr} = \{S \cap \{p_1, \ldots, p_n\} \mid S \in \Sscr\}$. The reduced set $\Tilde{\Sscr}$ is used to show the claims.

    Claims 1., 2.,3. and 4. are analogous. Hence, here we prove only Claim 1. Let $(y_1, y_2) \in M \times M$ such that for all $i \in \{1, \ldots, n\}$ $(y_1,y_2) \in p_i$ and let $p \in \Pcal$ such that $(y_1, y_2) \not\in p$. Let $S \in \Sscr \cap \{p_1, \ldots, p_n\}$ and take a closer look at (3) of Lemma \ref{th: restructuring Sscr}. Since $(y_1,y_2) \not\in p$, $S$ cannot be an element of the intersection of (3). Thus, $\{S\cap \{p_1, \ldots, p_n\} \in \Sscr \mid p\in\gamma(S)\}$ is empty and with the comment above we get that $D_n(p) = 0$.
\end{proof}

The next theorem gives an upper and lower bound on the cardinality of the elements $S \in \Sscr$.
\begin{theorem}\label{th: upper, lower bounds S}
For $\Sscr$, as defined in Section \ref{sec:ufg}, $\# S \ge 2$ and $\# S \le vc$ is true for all $S\in \Sscr$, where $vc$ is the VC dimension of the set $\gamma(2^{\Pcal})$.
\end{theorem}
\begin{proof}
Let $S \in \Sscr$. The proof for $\# S \ge 2$ follows immediately from Theorem \ref{th: S+card 2 and connect}.\\
To prove $\# S \le vc$ take an arbitrary subset $Q=\{p_1,\ldots, p_k\}\in \Sscr$ of size $k> vc$. Then this subset is not shatterable and thus there exists a subset $R\subseteq Q$ that cannot be obtained as an intersection of $Q$ and some $\gamma(S)$. In particular, with the extensivity of $\gamma$ it follows $\gamma(R) \cap Q \supsetneq R$ which means that there exists an order $p_i$ in $\gamma(R) \cap Q \backslash R$ for which the formal implication 
$R\rightarrow \{p_i\}$ holds. Thus, (because of the Armstrong rules, cf., \cite[p. 581]{armstrong74}) the order $p_i$ is redundant in the sense of $Q\backslash\{p_i\} \rightarrow Q$ and thus $Q$ is not minimal with respect to $\gamma$. Therefore, $Q$ is not in $\Sscr$ which completes the proof.
\end{proof}

Finally, we show the consistency of $D_n$.
\begin{theorem}\label{th: convergence}
    $D_n$ converges almost surely uniformly to $D$ for $n$ to infinity.
\end{theorem}
\begin{proof}
    Due to the i.i.d assumption and the law of large numbers, we know that for every $p \in \Pcal, \:\left\lVert \nu_n(p) - \nu(p)\right\rVert \overset{n \to \infty}{\to} 0$ almost surely (a.s). Since $\#\Pcal$ is finite, we get that $\nu_n$ also converges a.s. uniformly to $\nu$. Finally, we use that $D_n$ and $D$ are both the same finite composition of $\nu_n$ and $\nu$, respectively, and we obtain $
    \sup_{p \in \Pcal} \left\lVert D_n(p) - D(p)\right\rVert \overset{n \to \infty}{\to} 0 \text{ a.s.}$
\end{proof}
The last theorem states a contradiction to the claim that $D$ can be represented via pairwise comparisons. 
\begin{theorem}\label{thm_pairwise}
    $D_n$ cannot be represented as a function of the sum-statistics $w_{(a,b)}$.
\end{theorem}
\begin{proof}
    We simply give two data sets $\mathcal{D}=(p_1,p_2,p_3)$ and $\tilde{\mathcal{D}}=(\tilde{p}_1,\tilde{p}_2,\tilde{p}_3)$ on the basic set $M=\{y_1,y_2,y_3\}$ with the same sum-statistics but different associated depth functions: Let $p_1,p_2$ and $p_3$ be given as the transitive reflexive closures of $\{(y_1,y_2)\}$;  $\{(y_1,y_2),(y_1,y_3)\}$ and $\{(y_2,y_3), (y_1,y_3)\}$, respectively. Let  $\tilde{p}_1$, $\tilde{p}_2$ and $\tilde{p}_3$ be the transitive reflexive closure of $\{(y_1,y_2)\}$; $\{(y_1,y_3)\}$ and $\{(y_1,y_2),(y_2,y_3)\}$, respectively. Then both data sets have the same sum-statistics $w_{(y_1,y_2)}=w_{(y_1,y_3)}=2$; $w_{(y_1,y_3)}=1$ and $w_{(y_i,y_j)}=0$ for all other $y_i\neq y_j$. But the ufg depth of $p_1=\tilde{p}_1$ is $1/2$ w.r.t. the first data set but $7/10$ w.r.t the second data set. The corresponding code can be found at the link mentioned in Footnote~$1$.
\end{proof}


\acks{%
We thank all four anonymous reviewers for valuable comments that helped to improve the paper. Hannah Blocher and Georg Schollmeyer gratefully acknowledge the financial and general support of the LMU Mentoring program.
}

\section*{Author Contributions}
Hannah Blocher developed the idea of ufg depth. She wrote most of the paper. To be more precise:
The introduction was written by Christoph Jansen. Hannah Blocher wrote the preliminaries and defined the (empirical) ufg depth. Furthermore, Hannah Blocher claimed and proved Lemma~\ref{th: restructuring Sscr}, Theorem~\ref{th: S+card 2 and connect}, 1st and 2nd part, Corollary~\ref{cor: Dn always zero}, Corollary~\ref{cor: Dn and the sample}, the lower bound of Theorem~\ref{th: upper, lower bounds S} and Theorem~\ref{th: convergence}. Georg Schollmeyer made the claim and proofs of Theorem~\ref{th: S+card 2 and connect}, 3rd part and the upper bound of Theorem~\ref{th: upper, lower bounds S}. The claim of Theorem~\ref{thm_pairwise} was done by Georg Schollmeyer and Christoph Jansen. Georg Schollmeyer proved Theorem~\ref{thm_pairwise}. Chapter 5 was written by Georg Schollmeyer. Hannah Blocher wrote Chapter 6.1 and implemented the test if a subset is an element of $\Sscr$. Georg Schollmeyer contributed with the implementation of the connectedness property. Malte Nalenz provided the data set, performed the data preparation and wrote Chapter 6.2. Hannah Blocher analyzed the data set and wrote Chapter~6.3. Georg Schollmeyer, Christoph Jansen and Malte Nalenz supported the analysis with intensive discussions. Christoph Jansen provided the conclusion.

Georg Schollmeyer and Christoph Jansen also helped with discussions about the definition of ufg depth and all properties. Malte Nalenz, Christoph Jansen, and Georg Schollmeyer also contributed by providing detailed proofreading and help with the general structure of the paper.


\bibliography{reviewed_paper_version/blocher23}

\end{document}